\pdfoutput=1

\documentclass[letterpaper, 10 pt, conference]{ieeeconf}  % Comment this line out if you need a4paper

\IEEEoverridecommandlockouts                              % This command is only needed if 
\let\labelindent\relax

\usepackage{graphicx} % for pdf, bitmapped graphics files
\usepackage{epsfig} % for postscript graphics files
\usepackage{amsmath} 
\usepackage{amssymb} 
\usepackage{pifont}
\usepackage{soul,color} 
\usepackage{mathtools}
\usepackage{booktabs}
\usepackage{multirow}
\usepackage{enumerate}
\usepackage{balance}
\usepackage{gensymb}
\usepackage{tabulary}
\usepackage[numbers,sort&compress]{natbib}
\usepackage{dblfloatfix}
\usepackage{layouts}
\usepackage{algorithm}
\usepackage{algpseudocode}
\usepackage{lipsum}
\usepackage{subcaption}
\usepackage{enumitem}
\usepackage{siunitx}
\usepackage[bookmarks=false]{hyperref}
\usepackage{float}

\graphicspath{{figures/}}

\usepackage{subcaption}
\usepackage{tabulary}
\usepackage{multirow}

\usepackage{mwe}
\usepackage{xcolor}
\title{\LARGE \bf
CMU-GPR Dataset: Ground Penetrating Radar Dataset for \\ Robot Localization and Mapping \vspace{-1mm}}

\author{Alexander Baikovitz$^{1}$, Paloma Sodhi$^{1}$, Michael Dille$^{2}$, Michael Kaess$^{1}$ % <-this % stops a space
\thanks{\scriptsize The CMU-GPR dataset and related utility functions are available at \url{https://github.com/rpl-cmu/CMU-GPR-Dataset}. This work was supported by a NASA Space Technology Graduate Research Opportunity. }% <-this % stops a space
\thanks{\scriptsize $^{1}$ The Robotics Institute, Carnegie Mellon University, $^{2}$ Intelligent Robotics Group, KBR / NASA Ames Research Center, Correspondence to {\tt abaikovitz@cmu.edu}}
}%

\begin{document}

\maketitle

\thispagestyle{empty}
\pagestyle{empty}

%%%%%%%%%%%%%%%%%%%%%%%%%%%%%%%%%%%%%%%%%%%%%%%%%%%%%%%%%%%%%%%%%%%%%%%%%%%%%%%%
\begin{abstract}

% We present a novel dataset for research in subsurface perception captured using a custom experimental rig containing a ground penetrating radar (GPR). In total, the dataset contains 15 distinct trajectory sequences in 3 GPS-denied, indoor environments. Measurements from a GPR, wheel encoder, RGB camera, and inertial measurement unit were collected with ground truth positions from a robotic total station. In addition to the dataset, we provide a pipeline to produce processed images from the raw GPR data. This paper describes our recording platform, the data format, utility scripts, and proposed methods for using this data.

There has been exciting recent progress in using radar as a sensor for robot navigation due to its increased robustness to varying environmental conditions. However, within these different radar perception systems, ground penetrating radar (GPR) remains under-explored. By measuring structures beneath the ground, GPR can provide stable features that are less variant to ambient weather, scene, and lighting changes, making it a compelling choice for long-term spatio-temporal mapping. In this work, we present the CMU-GPR dataset---an open-source ground penetrating radar dataset for research in subsurface-aided perception for robot navigation. In total, the dataset contains 15 distinct trajectory sequences in 3 GPS-denied, indoor environments. Measurements from a GPR, wheel encoder, RGB camera, and inertial measurement unit were collected with ground truth positions from a robotic total station. In addition to the dataset, we also provide utility code to convert raw GPR data into processed images. This paper describes our recording platform, the data format, utility scripts, and proposed methods for using this data.

\end{abstract}

\section{Introduction}

We present the CMU-GPR dataset---the first open-source ground penetrating radar (GPR) dataset available to researchers interested in subsurface-aided perception for robot navigation, to the best of our knowledge. Radar-based perception has been shown to perform more robustly than conventional spatial or visual sensors in inclement weather~\cite{RadarRobotCarDatasetICRA2020, sheeny2020radiate, yspark-2019-icra-ws}. While recent work has focused on using millimeter-wave radar systems to construct surface-level models, there has been growing interest in using subsurface information from GPR for localization~\cite{Cornick2016, Ort2020}.

GPR presents a modality to recognize a location in situations where the visual environment may change by perceiving typically constant subsurface features. For instance, a robot operating in a mining environment may encounter substantial changes in the surface environment due to mining operations, yet underground features remain more consistent. Similarly, GPR-based localization can be effective in sparsely featured environments, such as monotonous tunnels and open roads, where subsurface geologic diversity can enable lane tracking with respect to a prior map.

In this contribution, we produce a dataset and utility functions to empower researchers to explore how GPR can be used as a tool for robot navigation applications. The dataset collected contains subsurface measurements from a low-cost, off-the-shelf, single-channel GPR system along with a wheel encoder, IMU, RGB camera, and robotic total station.

% GPR-based odometry has recently been shown to improve robot positioning in challenging indoor environments, which can be extended to enable robust positioning in dynamic and visually challenging settings.

\begin{figure}[!t]
    \centering
    \includegraphics[width=0.95\columnwidth]{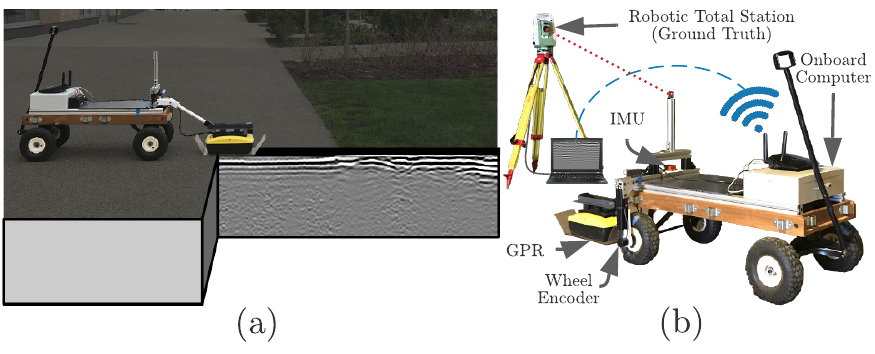}
    \caption{SuperVision platform for collecting subsurface data using GPR sensor during motion. (a) A cross section of the subsurface is shown containing different flat boundaries and a pipe. (b) Schematic of the SuperVision platform used for data collection.}
    \label{fig:cover}
    \vspace*{-.25in}
\end{figure}

\section{The SuperVision Platform}

SuperVision shown in Figure~\ref{fig:cover}(b) is a custom, manually-pulled experimental rig used for acquiring subsurface data from GPR. SuperVision uses a quad-core Intel NUC for compute and wirelessly transmits onboard data from an XSENS MTI-30 9-axis Inertial Measurement Unit, YUMO quadrature encoder with 1024 PPR, Intel RealSense D435 camera, and a Sensors and Software Noggin 500 GPR. Ground truth data was acquired by a Leica TS15 total station. The base station logs measurements from both the onboard computer and total station to ensure time synchronization.

\section{Data Collection}

The CMU-GPR dataset contains short trajectories from three distinct, GPS-denied environments: a basement (\textit{nsh\_b}), a factory floor (\textit{nsh\_h}), and a parking garage (\textit{gates\_g}). The dataset consists of 15 traversals where the manually-pulled experimental rig revisits previous locations, whether through forward-backward motion or by closing loops. Several distinct trajectories contain similar features, which are relevant to research on re-localization using subsurface information and further described on the project website. Additionally, we provide trajectories from outdoor environments that do not contain ground truth information, which present additional data to train models.

\begin{figure}
    \centering
    \includegraphics[width=\columnwidth]{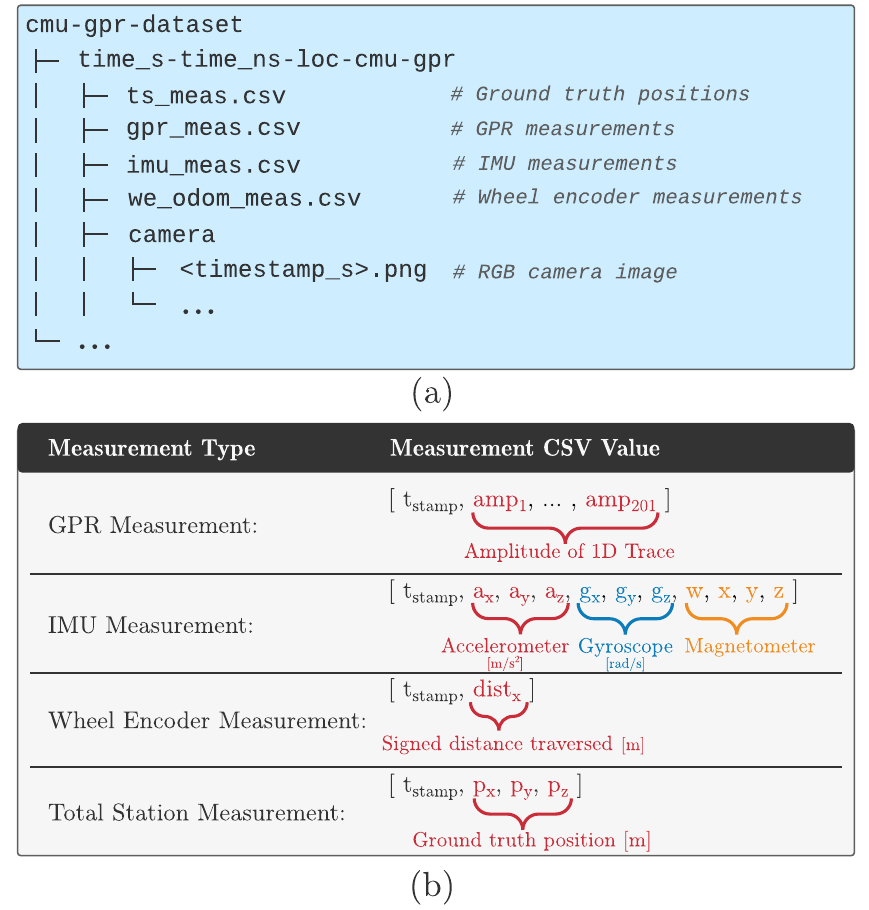}
    \caption{(a) Directory layout for the CMU-GPR dataset. (b) Data format by sensor measurement type.}
    \label{fig:directory_layout}
    \vspace*{-.2in}
\end{figure}

Figure~\ref{fig:directory_layout}(a) shows the directory structure for a single trajectory sequence. Each zip file contains data from a sequence, where each \texttt{.csv} represents a different measurement type. The maximum size of a zipped dataset file is 4 GB. The formats for each measurement type are summarized in Figure~\ref{fig:directory_layout}(b). In addition to the data, the project website contains relevant IMU noise and bias parameters as well as the factory extrinsic calibration of the system.

\section{Development Tools}
Along with relevant datasets for localizing GPR, we provide utility code, written in Python, which processes the raw GPR data to construct images. A modular routine to process 1D measurements and construct an image with uniform spacing is described in Section~\ref{signal_processing}. Additionally, we provide a script to generate submaps, which can be used for training or evaluating GPR sensor models.
\subsection{Signal Processing}
\label{signal_processing}

Utility code to process raw GPR signals and images is provided in \texttt{signal\_processing\_utils.py} and used in \texttt{metric\_gpr\_image.py}. The implementation accepts raw, unevenly spaced GPR images and produces processed brightness scans. The base pipeline performs rubber band interpolation, mean background subtraction, dewow filtering, triangular bandpass filtering, zero time correction, SEC gain, wavelet denoising, and gaussian filtering. 

\subsection{Image Construction}
\label{submap_construction}

Beyond processing the raw data, the utility code also simplifies image acquisition. The \texttt{MetricGprImage} object stores the GPR dataset and allows the client to access images based on the time of acquisition. The \texttt{ImageConstructor} object is even more abstract, allowing the client to create a traditional radargram of the entire sequence or automatically generate all valid submap images.

\begin{figure}
    \centering
    \includegraphics[width=0.95\columnwidth]{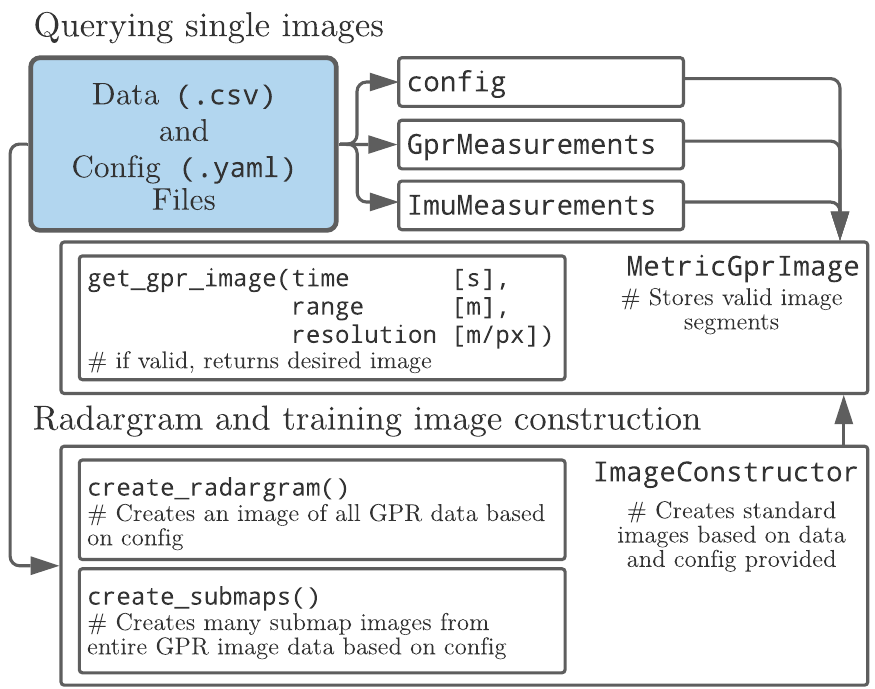}
    \caption{API for constructing processed images from CMU-GPR dataset.}
    \label{fig:my_label}
    \vspace*{-.2in}
\end{figure}

% \section{Proposed Method}
% This dataset was used to validate the submap-based approach, using the methods described in Section~\ref{submap_construction}, for localizing GPR demonstrated in~\cite{baikovitz2021ground}. In this approach, Baikovitz \textit{et al.} use a learned sensor model to use relative transformation predictions between GPR submaps to aid provided with the CMU-GPR dataset for localization.

\section{Approaches}
The dataset proposed in this paper was used for robot localization in an unknown, GPS-denied environments using a GPR sensor. In that work, Baikovitz \textit{et al.}~\cite{baikovitz2021ground} utilized learned sensor models to incorporate GPR submaps into a factor graph-based estimation framework. These learned models provided relative motion predictions during loop closures thereby correcting for accumulated drift in the position estimates. The goal of this work was to demonstrate how a low-cost, off-the-shelf, single-channel GPR system can be used effectively for robot localization.

We see many avenues for extending the use of this dataset and GPR in robotics. One limitation with a localization-only approach is that one needs to revisit prior locations in order to correct for drift. This can be avoided if we maintain an online map of the subsurface structures such as pipes, mines, and natural caves, and localize with respect to it. This would involve solving a simultaneous localization and mapping (SLAM) problem. Finding the correct representation for GPR data is likely the largest challenge for robust, GPR-based SLAM. We observe that a 2D image representation performs well compared to raw 1D traces, which do not provide unique enough features, and 3D point clouds, which are sensitive to noise and variable subsurface composition. Some prior work addresses automatic detection of hyperbolic features in 2D GPR images; however, these methods are often heuristic-based and not investigated for use in a SLAM graph~\cite{qingxuhyperbola2017}.

\section{Discussion}

Our motivation for providing this contribution is to encourage others in the field to take similar steps in making GPR-based perception datasets available to researchers. We believe that providing this data to the research community will spur further development of robust GPR-based localization systems for real world deployment.

\balance

\scriptsize
\bibliographystyle{IEEEtran}
\bibliography{main}

\begin{thebibliography}{1}
\providecommand{\url}[1]{#1}
\csname url@samestyle\endcsname
\providecommand{\newblock}{\relax}
\providecommand{\bibinfo}[2]{#2}
\providecommand{\BIBentrySTDinterwordspacing}{\spaceskip=0pt\relax}
\providecommand{\BIBentryALTinterwordstretchfactor}{4}
\providecommand{\BIBentryALTinterwordspacing}{\spaceskip=\fontdimen2\font plus
\BIBentryALTinterwordstretchfactor\fontdimen3\font minus
  \fontdimen4\font\relax}
\providecommand{\BIBforeignlanguage}[2]{{%
\expandafter\ifx\csname l@#1\endcsname\relax
\typeout{** WARNING: IEEEtran.bst: No hyphenation pattern has been}%
\typeout{** loaded for the language `#1'. Using the pattern for}%
\typeout{** the default language instead.}%
\else
\language=\csname l@#1\endcsname
\fi
#2}}
\providecommand{\BIBdecl}{\relax}
\BIBdecl

\bibitem{RadarRobotCarDatasetICRA2020}
D.~Barnes, M.~Gadd, P.~Murcutt, P.~Newman, and I.~Posner, ``The oxford radar
  robotcar dataset: A radar extension to the oxford robotcar dataset,'' in
  \emph{IEEE Intl. Conf. on Robotics and Automation (ICRA)}, Paris, 2020.

\bibitem{sheeny2020radiate}
M.~Sheeny, E.~De~Pellegrin, S.~Mukherjee, A.~Ahrabian, S.~Wang, and A.~Wallace,
  ``Radiate: A radar dataset for automotive perception,'' \emph{arXiv}, 2020.

\bibitem{yspark-2019-icra-ws}
Y.~S. Park, J.~Jeong, Y.~Shin, and A.~Kim, ``Radar dataset for robust
  localization and mapping in urban environment,'' in \emph{ICRA Workshop on
  Dataset Generation and Benchmarking of SLAM Algorithms for Robotics and
  VR/AR}, 2019.

\bibitem{Cornick2016}
M.~Cornick, J.~Koechling, B.~Stanley, and B.~Zhang, ``{Localizing Ground
  Penetrating RADAR: A Step Toward Robust Autonomous Ground Vehicle
  Localization},'' \emph{Journal of Field Robotics}, vol.~33, no.~1, pp.
  82--102, jan 2016.

\bibitem{Ort2020}
T.~Ort, I.~Gilitschenski, and D.~Rus, ``{Autonomous navigation in inclement
  weather based on a localizing ground penetrating radar},'' \emph{IEEE
  Robotics and Automation Letters}, vol.~5, no.~2, pp. 3267--3274, 2020.

\bibitem{baikovitz2021ground}
A.~Baikovitz, P.~Sodhi, M.~Dille, and M.~Kaess, ``Ground encoding: Learned
  factor graph-based models for localizing ground penetrating radar,'' in
  \emph{arXiv}, 2021.

\bibitem{qingxuhyperbola2017}
Q.~Dou, L.~Wei, D.~R. Magee, and A.~G. Cohn, in \emph{Real-Time Hyperbola
  Recognition and Fitting in GPR Data}, 2017.

\end{thebibliography}

\end{document}